\pdfoutput=1
\documentclass{article}
\usepackage{ijcai20}

\usepackage{times}
\usepackage{soul}
\usepackage{url}
\usepackage[hidelinks]{hyperref}
\usepackage[utf8]{inputenc}
\usepackage[small]{caption}
\usepackage{graphicx}
\usepackage{amsmath}
\usepackage{amsthm}
\usepackage{booktabs}
\usepackage{algorithm}
\usepackage{algorithmic}
\usepackage{bm}
\usepackage{subcaption}

\newcommand{\R}{\mathcal R}

\newcommand{\T}{\mathrm T}

\newcommand{\ve}[1]{\bm#1}  
\newcommand{\m}[1]{\bm#1}  

\usepackage{enumitem}
\setitemize{noitemsep,nosep}

\usepackage{placeins}

\title{Pixel Embedding: Fully Quantized Convolutional Neural Network with Differentiable Lookup Table}

\author{
Hiroyuki~Tokunaga
\and
Joel~Nicholls\and
Daria~Vazhenina\footnote{Currently with Woven by Toyota, Inc., Tokyo, Japan.}\and
Atsunori~Kanemura\footnote{Currently with Mirai Hoshu K.K., Tsukuba, Japan.}
\affiliations
LeapMind Inc., Tokyo 150-0044, Japan\\
\emails
tokunaga.hiroyuki@gmail.com, joelowennicholls@gmail.com, daria.vazhenina@woven.toyota, atsu-kan@mirai-hoshu.co.jp
}

\begin{document}

\maketitle

\begin{abstract}
By quantizing network weights and activations to low bitwidth, we can obtain hardware-friendly and energy-efficient networks.
However, existing quantization techniques utilizing the straight-through estimator and piecewise constant functions face the issue of how to represent originally high-bit input data with low-bit values.
%
To fully quantize deep neural networks, we propose pixel embedding, which replaces each float-valued input pixel with a vector of quantized values by using a lookup table. The lookup table or low-bit representation of pixels is differentiable and trainable by backpropagation. Such replacement of inputs with vectors is similar to word embedding in the natural language processing field.
%
Experiments on ImageNet and CIFAR-100 show that pixel embedding reduces the top-5 error gap caused by quantizing the floating points at the first layer to only 1\% for the ImageNet dataset, and the top-1 error gap caused by quantizing first and last layers to slightly over 1\% for the CIFAR-100 dataset. The usefulness of pixel embedding is further demonstrated by inference time measurements, which demonstrate over 1.7 times speedup compared to floating point precision first layer.
\end{abstract}

\section{Introduction}

After achieving record-breaking performance in various tasks represented by image classification~\cite{krizhevsky2012imagenet,resnet}, deep learning is expected to operate in every aspect of our life. Reducing neural network footprint to fit into embedded devices enables use of deep learning techniques in e.g.\ wearable devices, drones, or vehicles, which are often isolated from the network and have limited computational resources.

This paper focuses on quantization as a solution for accelerating inference in neural networks for embedded, on-device applications~\cite{guo2017survey}. Quantized neural networks utilize low-bit parameters and operations, greatly reducing the model size and computation necessary for inference.

With FPGAs or customized ASICs, low-bit operations can be efficiently implemented in logical circuits with bit operations. A 32 bit floating point multiplication unit consumes more than 50 times more logic elements than 4 bit integer~\cite{guo2017survey}. This means that the theoretical speed up factor of quantized computation for FPGAs or customized ASICs is over 50, if memory bandwidth is not a bottleneck.

Quantizing the first convolutional layer is one of the largest technical issues for accelerating quantized neural networks. Indeed, work by Liu {\em et al.}~\shortcite{liu2019concrete} shows that the first layer inputs are typically less resilient to quantization compared to the other layers.

Most works on neural network quantization keep the first layer as non-quantized to avoid severe accuracy drops~\cite{binarynet,xnornetwork}. However, utilizing high precision for the first layer is not an effective solution. In the case of FPGAs or ASICs, the first non-quantized convolutional layer is a huge performance bottleneck, since the intermediate (quantized) layers have been accelerated. We found that the non-quantized first convolutional layer of the ResNet-18 classification network consumes 54\% of wall clock time on our FPGA-based neural network accelerator.

In this paper, we propose a technique we term {\em pixel embedding} to quantize both weights and activations of the first convolutional layer of deep convolutional neural networks (CNNs), improving the accuracy loss compared to the existing quantization methods. The pixel embedding method extends an idea from word embeddings in natural language processing~\cite{nnlm} to quantization.

Pixel embedding replaces input data with a more easily quantizable form (Section~\ref{sec:pe}). The resulting operation at inference time is a lookup table of low-bit values. Experiments on CIFAR-100 and ImageNet show that pixel embedding achieves state-of-the-art for quantizing the first layer compared to the existing quantization methods  (Section~\ref{sec:exp}).
Our main contributions are:
\begin{itemize}
    \item With the pixel embedding method, we show that it is possible to quantize both weights and input data for the first convolutional layer, breaking a general belief that quantizing the first or last layers is difficult.
    \item We examine the strength of the pixel embedding method through experiments on two widely used image classification datasets, CIFAR-100 and ImageNet, confirming accuracy improvement over existing quantization methods. Specifically, on ImageNet, the top-5 accuracy is almost 10\% higher than just quantizing input data. For CIFAR-100, we achieve 4.4\% increase in top-1 accuracy over simple quantization of input data.
    \item We further validate the usefulness of pixel embedding by measuring inference time speedup on our FPGA-based neural network accelerator.
\end{itemize}

\section{Existing quantization techniques}
\label{sec:rw}

Convolutional neural networks are computationally heavy algorithms. It is desirable to reduce energy consumption and memory requirement by executing computation in lower precision like 8 bit integer~\cite{krishnamoorthi2018quantizing}, instead of 32 bit floating point. Such low-bit approximation is called {\em quantization}~\cite{qnnHubara}. Taking this to the extreme, 1 bit integer can be employed~\cite{binarynet,xnornetwork}. This paper is concerned with low bitwidth quantization, which can be used for acceleration of neural networks on dedicated hardware.

Quantization-aware training uses a combination of techniques that allows for very low-precision convolutions, while keeping high accuracy~\cite{qnnHubara,dorefaNetZhou}. To learn quantized weight values during training, a quantization operator is used on the high bitwidth weights in the forward pass of training. For backpropagation, the straight-through estimator (STE) method~\cite{bengio2013ste,binarynet} can be used. An STE can be thought of as an operator that has mismatched forward and backward operations.

In addition to weights, the activations of convolutional layers can also be quantized during training, and this quantization operation remains in the inference stage~\cite{binarynet}, unlike the weight quantization operation. Piece-wise constant functions are used for the quantization of activations. These operations have low computation cost and have the property that similar activations get mapped to the same quantized values, preserving the notion of closeness.

The combination of both weight and activation quantization allows for convolution to be performed using low-bitwidth operations, giving a great boost to the algorithm's speed and efficiency. Although the first layer is typically relatively small compared to the neural network as a whole, using a floating point first layer becomes a serious bottleneck when the other layers have been optimized using quantization. Indeed, the first floating point layer takes up 54\% of the inference time in our implementation.

Although most previous works have used fully floating point precision for the first layer convolution, there are exceptions. In Zhou {\em et al.}~\shortcite{dorefaNetZhou}, a network with binary weights and 8-bit inputs is implemented and evaluated. Using binary weights eliminates multiplications, reducing the convolution operation to integer additions.

In the work of D{\"u}richen {\em et al.}~\shortcite{binaryinput}, each integer input is separated bit-wise into separate channels and convolution is performed between those binary activations and binary weights (called direct binary input data), then an additional $1\times1$ convolution is used (called binary input layer). The binary input layer result for classification on the CIFAR-10 dataset incurs a loss of 4.58\% accuracy; however, impressive results were achieved for the physical activity monitoring (PAMAP2) dataset, actually decreasing validation error by 1.92\%.

\section{Fully quantized networks}
\label{sec:pe}

Pixel embedding is inspired by word embedding in the natural language processing field~\cite{nnlm}, where each word is associated with a real-valued vector. Embeddings are powerful because they can be learned by backpropagation. To understand this, we need to introduce the concepts of 1-hot representations and embedding lookup tables. Using these two concepts, we then describe pixel embedding.

\subsection{1-hot Representation}
\label{subsec:1hotrepresentation}

Let $w \in \mathcal V$ be a word id, where $\mathcal V = \{0, 1, \dotsc, N-1\}$ is a vocabulary comprising all ids of unique words.
The set of $N$-length 1-hot vectors is
\begin{equation}
\mathcal H_N = \Bigl\{\bm h \in \{0, 1\}^N \Bigm| \sum_{n=1}^N h_n = 1\Bigr\}.
\end{equation}
That is, only one element is $1$ and the other elements are $0$.

Let $\bm h\colon \mathcal V \to \mathcal H_N$ be a vector-valued function that maps a word $w \in \mathcal V$ to its 1-hot representation. That is, $\bm h(w) \in \mathcal H_N$ and $h_n(w) = 1$ if $n = w$ and $h_n(w) = 0$ if $n \neq w$.
For example, $\bm h(1) = [0, 1, 0, \dotsc]^\T$ (note that indexing is 0-based).

\subsection{Embedding Lookup Table}
\label{subsec:embeddinglookuptable}

We denote the embedding of $w \in \mathcal V$ by $\ve e_w \in \R^d$, where $d > 0$ is the dimensionality of the embedding space and $\R$ is the set of real numbers.
We can formulate the conversion process as a matrix multiplication. Let
\begin{equation}
  \m E = \begin{bmatrix}
           | & | &  & |\\
           \ve e_0 & \ve e_1 & \dotsc & \ve e_{N-1}\\
           | & | &  & |
         \end{bmatrix} \in \R^{d \times N}
\end{equation}
be a matrix whose columns are embedding vectors.
The matrix $\m E$ is called an {\em embedding lookup table}.
Then, using the 1-hot representation, we have
\begin{equation}
  \ve e_w = \m E \ve h(w).
\end{equation}

The matrix-vector formulation allows us to learn embeddings through backpropagation, because ``choosing a column from a matrix'' becomes a differentiable operation. Several word embedding algorithms made use of the fact that embedding lookup tables can be optimised with backpropagation to learn embeddings~\cite{nnlm,word2vec,glove}.

\subsection{Pixel Embedding}
\label{subsec:pixelembedding}

\begin{figure}
  \small
  \hfil
  \begin{minipage}{0.49\columnwidth}\centering
  \begin{tabular}{r@{$\:$}c@{$\:$}l}
    \toprule
    $P$ & & Embedding $\ve e$ \\
    \midrule
    $0$& $\mapsto$ & $(3301)_\text{quaternary}$ \\
    $1$& $\mapsto$ & $(1230)_\text{quaternary}$ \\
    & $\vdots$ & \\
    $255$& $\mapsto$&  $(0112)_\text{quaternary}$ \\
    \bottomrule
  \end{tabular}
  \subcaption{$d = 4$, $Q = 2$}
  \end{minipage}
\hfil
  \begin{minipage}{0.49\columnwidth}\centering
  \begin{tabular}{r@{$\:$}c@{$\:$}l}
    \toprule
    $P$ & & Embedding $\ve e$ \\
    \midrule
    $0$& $\mapsto$ & $(01010)_\text{binary}$ \\
    $1$& $\mapsto$ & $(11001)_\text{binary}$ \\
    & $\vdots$ & \\
    $255$& $\mapsto$&  $(01001)_\text{binary}$ \\
    \bottomrule
  \end{tabular}
  \subcaption{$d = 5$, $Q = 1$}
  \end{minipage}
  \caption{Examples of pixel embedding (for one color component). An 8-bit integer $P$ representing R, G, or B is replaced with a low-bitwidth embedding $\ve e$ in $\mathcal L_Q^d$.}
  \label{fig:embeddingLookupTable}
\end{figure}

Although an image pixel is usually represented by $[P_\mathrm R,P_\mathrm G,P_\mathrm B] \in \{0, 1, \dotsc, 255\}^3$, where each of the R, G, B channels has 8 bits, we can employ 1-hot representation to express a pixel by $[\ve h_R, \ve h_G, \ve h_B] \in \mathcal H_{256}^3 \subset \{0, 1\}^{256 \times 3}$, giving a $256$-length 1-hot vector per color component. By using the $\ve h(\cdot)$ function defined in Section~\ref{sec:pe}, we can write the 1-hot representation to be $[\bm h(P_\mathrm R), \bm h(P_\mathrm G), \bm h(P_\mathrm B)]$.
The combination of 1-hot representation and embedding makes it possible to replace input 8-bit color components with an easily quantizable representation.

Since we would like to have low-bit representations, we define our embedding lookup table $\m E$ to take its values not in $\R$ but in $\mathcal L_Q$, which is the set of $Q$-bit numbers. We select $Q$ to be the desired bitwidth of quantization; $Q = 1, 2, 3$ means binary, quaternary, and octal activations, respectively.

The embedding of a pixel $[P_\mathrm R, P_\mathrm G, P_\mathrm B]$ using the embedding lookup table $\m E \in \mathcal L_Q^{d \times 256}$ is defined to be the product of $\m E$ and 1-hot pixel $[\ve h(P_\mathrm R), \ve h(P_\mathrm G), \ve h(P_\mathrm B)]$:
\begin{align}
    \ve e &= \m E [\ve h(P_\mathrm R), \ve h(P_\mathrm G), \ve h(P_\mathrm B)]\\
    &= [\m E\ve h(P_\mathrm R), \m E\ve h(P_\mathrm G), \m E\ve h(P_\mathrm B)] \in \mathcal L_Q^{d \times 3}.
\label{eq:pixelembedding}
\end{align}%
Each color component of the input image is encoded as a $d$-dimensional vector of $Q$-bit numbers. The channel dimensions are therefore expanded $d$ times, resulting in $3d$ feature maps quantized to $Q$ bits, which can be used as input for the subsequent convolutional layers. Examples of pixel embeddings are shown in Figure~\ref{fig:embeddingLookupTable}.

For training the embedding lookup table, we use an embedding that maps to floating point precision, followed by quantization function (and STE for the backpropagation~\cite{bengio2013ste}). At the inference stage, the floating point precision lookup table and quantization function are merged, resulting in an embedding lookup table that maps to $Q$-bit numbers. This greatly reduces the memory overhead required for inference. When the input is an 8 bit number and the output is a $d$-dimensional vector of $Q$-bit numbers, the entire lookup table contains only $256 d Q$ bits in total.

\subsection{Quantizing the last layer}
\label{subsec:lastlayer}

For the task of classification, quantization of the final layer is fairly straightforward. The immediate output of a quantized layer is integer-valued. The integer outputs of a final quantized layer can be fed directly to the loss function for network training. Note that for the inference stage, the softmax function can be removed and integer comparison can be used to give the class prediction.

Object detection requires a little trick to be able to implement a quantized final layer while also keeping good accuracy. The outputs of the final layer convolution are used for box predictions. However, the outputs of a quantized convolution must be integer, and this will generally not be able to give the correct scaling for box predictions. Therefore, scaling the outputs by a floating point precision multiplication can be used to get the correct scale. Using a single factor for scaling is sufficient, which only introduces a very small computational overhead.

\section{Experiments}
\label{sec:exp}

\begin{table*}[t]
\centering\small
\begin{tabular}{lccc}
\toprule
 & \multicolumn{2}{c}{ImageNet} & CIFAR-100 \\
\cmidrule(lr){2-3}\cmidrule(lr){4-4}
Method & Top-1 acc. & Top-5 acc. & Top-1 acc. \\
\midrule
Weight quantization & 51.22\% & 75.90\%  & 63.73\% \\
Input and weight quantization & 43.72\% & 68.68\% & 61.47\% \\
Pixel embedding (ours) & 54.51\%  & 78.51\% & 65.87\% \\
\midrule
First layer full precision & 55.85\% & 79.69\% & 66.93\% \\
\bottomrule
\end{tabular}
\caption{Single-crop top-1 and top-5 validation accuracy for ImageNet and CIFAR-100 classification tasks. Note that our method quantizes both inputs and weights.}
\label{table:exp1}
\end{table*}

To measure the accuracy of the proposed method compared to baseline methods, we perform experiments on ImageNet and CIFAR-100 datasets. To measure speed, we run inference on imagenet-size images, using our FPGA-based neural network accelerator. For the ImageNet experiments, the first and intermediate layers are quantized. For the CIFAR-100 experiments, all layers are quantized, including the last layer.

The quantization bitwidth used in this paper is 2 and 1 bits for activations and weights, respectively. For standard quantization layers, activations are quantized via uniform quantization and weights are quantized using the sign function and scaling factors.

When using pixel embedding for the first layer, pixel embedding quantizes the inputs to 2 bits, and weights use standard quantization. Imagenet experiments use embedding dimension $d = 16$ and CIFAR-100 experiments use $d=8$.

We compare pixel embedding with the following three methods of dealing with the first layer:
\begin{enumerate}[noitemsep,nosep]
  \item Weight quantization: The weights are quantized to 1 bit.
  \item Input and weight quantization: The input pixel values are quantized to 2 bits and the weights to 1 bit.
  \item First layer full precision: Both the weights and activations are kept at full precision. This is the baseline for when the first layer is not quantized.
\end{enumerate}

We evaluated our method on the ImageNet Large Scale Visual Recognition Challenge 2012 (ILSVRC2012) classification dataset~\cite{krizhevsky2012imagenet}, to check the robustness of our method. The models were trained on the 1.28 million training images, and evaluated on the 50k validation images. We adopted an 18-layer residual network as the base network, and compared top-1 accuracy and top-5 accuracy with different quantization settings. We followed most of the training settings and hyper-parameters used in~\cite{resnet}, with two exceptions: (i)~we used a batch size of 160; (ii)~we decreased the learning rate by a factor of 10 at 100k, 200k, and 250k steps. Model training took on the order of one or two days for 300k iterations on 8 GPUs.

We further verify our findings by evaluating on the CIFAR-100 dataset~\cite{cifar100}, which consists of 50k training images and 10k testing images in 100 classes. Each image is 32 by 32 pixels, and we used only basic data augmentation---pad, crop, and horizontal reflection. The top-1 accuracy was used to measure performance. For these experiments, we used two less downsampling operations compared to the usual ResNet-18 version of He {\em et al.}~\cite{resnet}, to better suit CIFAR-size images.

\subsection{Accuracy measurements}
\label{subsec:accuracy}

\begin{figure*}[th]
\centering
\includegraphics[width=0.8\linewidth]{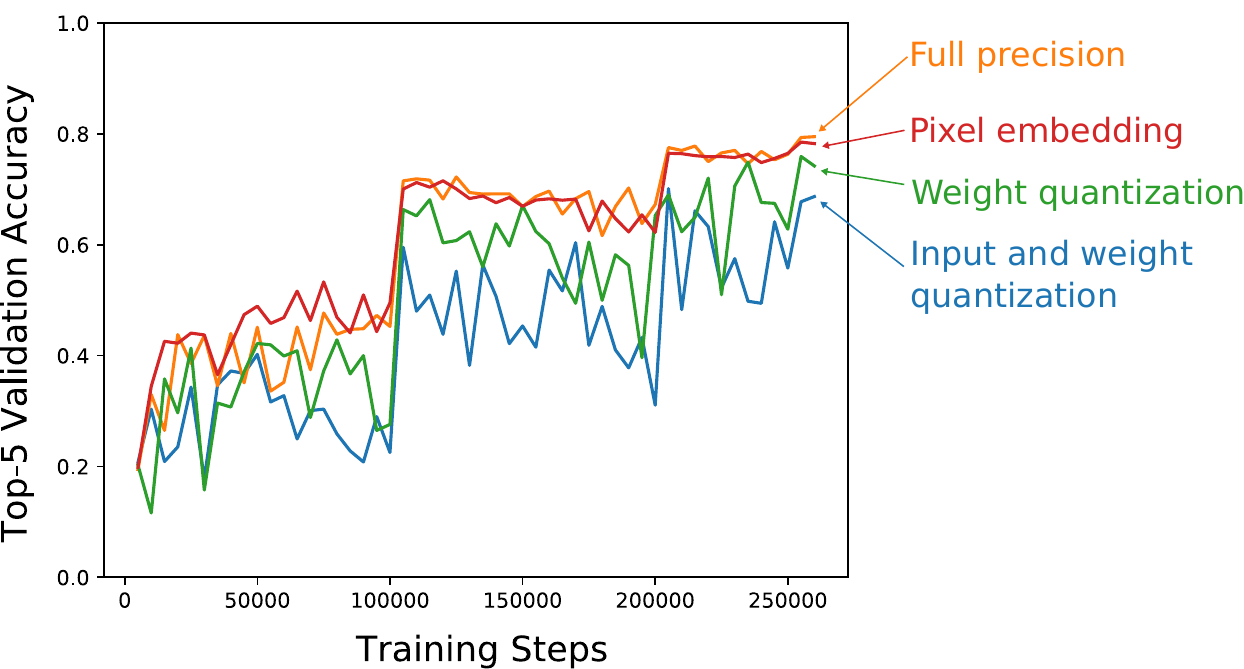}
\caption{The validation curves of the four methods on the ImageNet dataset. Full precision refers to the network having full precision first layer and quantized intermediate layers. These plots show that pixel embedding is not just superior to naive quantization, but also convergence is more stable.}\label{fig:exp1}
\end{figure*}

First, we evaluate pixel embedding in terms of classification accuracy using ImageNet and CIFAR-100 datasets.

Table~\ref{table:exp1} compares the top-1 and top-5 accuracies of pixel embedding and the other methods. For both datasets, we found a similar trend in results. Pixel embedding drops only a small amount of accuracy compared to floating point precision first layer, and achieves higher accuracy compared to the two standard methods of first layer quantization. The significant accuracy drop due to standard quantization of the first layer is consistent with results reported in \cite{binaryinput}.

For ImageNet, standard quantization of the inputs and weights of the first layer resulted in 43.72\% and 68.68\% top-1 and top-5 accuracy. Instead, using pixel embedding for quantization of the first layer resulted in 54.51\% and 78.51\% top-1 and top-5 accuracy. This reduces the gap to the floating point precision first layer significantly, in both top-1 and top-5 accuracy.

For CIFAR-100, top-1 accuracy was measured. The performance gap between floating point precision first layer and the standard input and weight quantization was 5.45\%. When instead using pixel embedding, the performance gap was reduced to 1.05\%. This result is in qualitative agreement with the Imagenet results.

Figure~\ref{fig:exp1} shows the validation curves of the top-5 validation accuracies at each 1k iterations on the ImageNet dataset. The curve of the model with pixel embedding is smoother than the other two quantized models, suggesting the robustness of pixel embedding against the existing quantization methods.

The validation curve for weight quantization shows that only quantizing the weights of the first layer can improve accuracy compared to the quantization of both inputs and weights. However, the validation curves of both are lower and fluctuate more heavily than the pixel embedding curve. We emphasize that our method achieved very small error gap to the full precision setting; the proposed method is a robust solution to complicated real problems.

\subsection{Inference time measurements}
\label{subsec:inferencetimes}

We evaluate pixel embedding via inference time measurements, using the same model settings as described above for the ResNet-18 classifier on Imagenet. In particular, the last layer uses floating point precision and pixel embedding uses embedding dimension 16. The image size is 224 $\times$ 224 after nearest neighbour downsampling, which is not included in the inference time measurement. This set of experiments was performed using Intel Cyclone V FPGA embedded in Terasic DE10-Nano development kit. It is accompanied by a dual-core ARM A9 processor running at 800 MHz with 64KB RAM.

The results of inference experiments are shown in Table~\ref{table:inferencetimes}. The mean and standard deviation of each of these results were obtained by 20 separate runs on a single image. The fastest network is the one using input and weight quantization ($158.2 \pm 0.3$ ms). Pixel embedding is only slightly slower ($160.8 \pm 0.3$ ms), and first layer full precision takes much longer ($279.4 \pm 0.7$ ms). Pixel embedding achieves 1.7 times speedup compared to first layer full precision.

\begin{figure}[t]
\centering
\includegraphics[width=\linewidth]{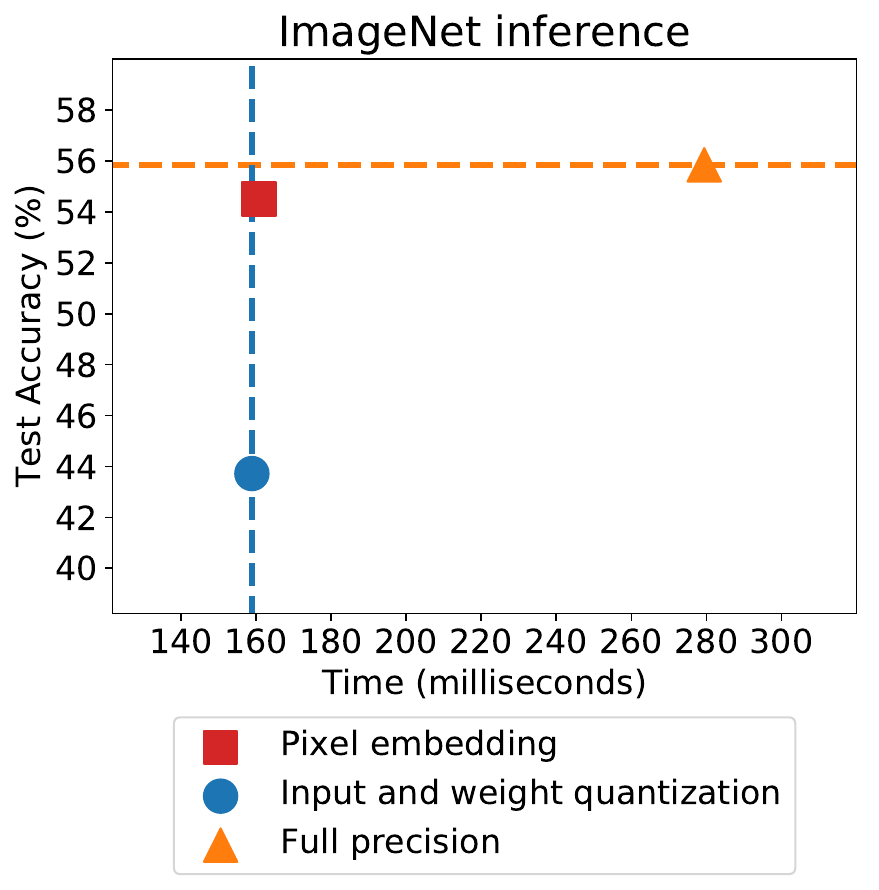}
\caption{Scatter plot of models, evaluated on the two objectives of test accuracy versus inference time. All models use ResNet-18 for ImageNet classification. The vertical dashed line indicates the inference time of the fastest model (input and weight quantization). The horizontal dashed line indicates the test accuracy of the most accurate model (full precision). Pixel embedding is close to the intersection of these dashed lines, so it is strong in both objectives.}\label{fig:tradeoff}
\end{figure}

Taking into consideration the combined objectives of accuracy and inference time, none of the three methods strictly dominate any of the others in terms of pareto optimality. However, pixel embedding is close to both fastest and most accurate. This point is illustrated more clearly in Figure~\ref{fig:tradeoff}. In the trade-off between accuracy and speed, pixel embedding is the best practical choice.

\begin{table}
\centering\small
\begin{tabular}{lccc}
\toprule
Method & Mean time (ms) \\
\midrule
Input and weight quantization & $158.9 \pm 0.3$ \\
Pixel embedding (ours) & $160.8 \pm 0.3$ \\
\midrule
First layer full precision & $279.4 \pm 0.7$ \\
\bottomrule
\end{tabular}
\caption{Inference time measurements for the ImageNet classification task on our FPGA-based neural network accelerator. When running inference, we alternate between the three methods 20 times. For each method, the mean and standard deviation are calculated using those 20 samples. These time measurements do not include preprocessing (nearest neighbour downsampling).}
\label{table:inferencetimes}
\end{table}

\section{Conclusion}
\label{sec:conclusion}

In this paper, we have shown that it is possible to quantize both the input data and weights for deep convolutional neural networks. We proposed {\em pixel embedding}, which converts input data into more easily quantizable form. Furthermore, we conducted experiments that revealed the clear advantage of the proposed approach. The top-5 accuracy for ImageNet was almost 10\% higher than just quantizing input data and weight. We found similar results for the CIFAR-100 dataset, where pixel embedding increases accuracy by 4.4\% compared to standard quantization of the first layer.

Using pixel embedding quantization and floating point values in the last layer, we can reduce the increase of top-1 error due to first layer quantization to just 1.34\% for the ImageNet dataset. In addition to accuracy, we also measured the inference time of classification with pixel embedding, which was over 1.7 times quicker than the floating point precision alternative. Because of this, pixel embedding enables both fast and accurate inference.

Although we have obtained valuable insights into the new problem, there is a possibility to import more ideas from other research fields like natural language processing. For example, it might be possible to obtain better embeddings by employing recently proposed deep compositional code learning~\cite{shu2018compressing}.

\FloatBarrier

\bibliographystyle{named}
\bibliography{pixembbib}

\end{document}